\pdfoutput=1
\documentclass[11pt]{article}

\usepackage{ACL2023}

\usepackage{times}
\usepackage{latexsym}

\usepackage[T1]{fontenc}
\usepackage[utf8]{inputenc}

\usepackage{microtype}

\usepackage{inconsolata}

\usepackage{booktabs}
\usepackage{todonotes}

\usepackage{pgfplots}
\pgfplotsset{compat=1.18}
\usepackage{tikz}

\title{Large Language Models and Prompt Engineering for Biomedical Query Focused Multi-Document Summarisation}

\author{Diego Moll\'a \\
  School of Computing \\
  Macquarie University \\
  Sydney, Australia \\
  \texttt{diego.molla-aliod@mq.edu.au}}

\begin{document}
\maketitle
\begin{abstract}
This paper reports on the use of prompt engineering and GPT-3.5 for biomedical query-focused multi-document summarisation. Using GPT-3.5 and appropriate prompts, our system achieves top ROUGE-F1 results in the task of obtaining short-paragraph-sized answers to biomedical questions in the 2023 BioASQ Challenge (BioASQ 11b). This paper confirms what has been observed in other domains: 1) Prompts that incorporated few-shot samples generally improved on their counterpart zero-shot variants; 2) The largest improvement was achieved by retrieval augmented generation. The fact that these prompts allow our top runs to rank within the top two runs of BioASQ 11b demonstrate the power of using adequate prompts for Large Language Models in general, and GPT-3.5 in particular, for query-focused summarisation.
\end{abstract}

\section{Introduction}

% Introduction text based on ChatGPT output

Large Language Models (LLMs), such as GPT-3.5, have gained significant attention in recent years for their ability to generate coherent and contextually relevant text. These models have been useful for various Natural Language Processing (NLP) tasks, including text classification, text summarisation, and question answering (QA).

Biomedical QA involves the generation of accurate and concise answers to questions based on information available in biomedical publications such as PubMed. This can be seen as a task of query-focused multi-document summarisation since the answers are generated from the text of one or more publications and, in contrast to ``factoid'' or ``multiple-choice QA'', they need to be presented as stand-alone text containing one or several sentences. The biomedical domain poses unique challenges due to the domain-specific terminology, and the need for precise and contextually appropriate answers.

In this paper, we present our approach to biomedical query-focused multi-document summarisation using GPT-3.5 and prompt engineering techniques. Specifically, we investigate the impact of using few-shot and retrieval-augmented generation (RAG). Our approach ranks among the two top results in the task of obtaining short-paragraph-sized answers to biomedical questions, as evaluated in the 2023 BioASQ Challenge (BioASQ 11b).\footnote{Code associated with this paper is publicly available at \url{https://github.com/dmollaaliod/bioasq11b-public}.}

\section{Background and Related Work}\label{sec:related}

LLMs' tendency to ``hallucinate'' and generate text that does not align to the user intention has been documented widely. The main reason for this is the wide range of text types used during the pre-training stage of these LLMs. Given the highly technical nature of the biomedical domain, and the fact that a relatively small percentage of texts used during the pre-training stage is within the biomedical domain, this problem is likely to be exacerbated in such domain. A common solution for QA is to use RAG approaches by incorporating relevant snippets in the prompt. For example, for multiple-choice medical QA, \citet{Lievin2023} used GPT-3.5 and injected passages of Wikipedia into the prompt. They experimented with the use of various prompt strategies such as Chain-of-Thought, zero-shot, and few-shot.

BioASQ\footnote{\url{http://bioasq.org/}} organises challenges on various tasks related to biomedical semantic indexing and QA \cite{Krithara2023}. Of interest for this paper is Task~B, which evaluates the two main phases of open-ended QA systems. In Phase~A, systems are expected to retrieve relevant snippets from the PubMed biomedical research papers repository. Then, in Phase~B, systems are expected to give the specific answers to the biomedical questions. There are four types of questions: ``yes/no'', ``factoid'', ''list'' and ``summary'', and two types of answers are possible: ``exact'' answers, where the answers are provided without relevant context, and ``ideal'' answers, where the answers are short paragraphs. Figure~\ref{fig:example} shows an example of an exact and an ideal answer to a ``factoid'' question from the BioASQ data.
\begin{figure}
    \centering
\fbox{
\begin{minipage}{7cm}
\em
    \begin{description}
      \item[Question] Which disease phenotype has the worst prognosis in Duchenne Muscular Dystrophy?
      \item[Exact answer] Dp140 isoform
      \item[Ideal answer] Dp140 isoform is related to increased risk of cognitive impairment and thus worse prognosis.
    \end{description}
\end{minipage}
}
    \caption{A factoid question from the BioASQ data sets, together with its ``exact'' and its ``ideal'' answer.}
    \label{fig:example}
\end{figure}
Table~\ref{tab:bioasqstats} shows some statistics about the dataset used in BioASQ 11b.
\begin{table}[]
    \centering
    \begin{tabular}{lrrrrr}
    \toprule
    Partition     & y/n & fact & list & sum & total\\
    \midrule
    Training     & 1271 & 1417 & 901 & 1130 & 4719\\
    Batch 1 & 24 & 19 & 12 & 20 & 75\\
    Batch 2 & 24 & 22 & 12 & 17 & 75\\
    Batch 3 & 24 & 26 & 18 & 22 & 90\\
    Batch 4 & 14 & 31 & 24 & 21 & 90\\
    \bottomrule
    \end{tabular}
    \caption{Statistics of the BioASQ 11b dataset. Batches 1 to 4 are test batches.}
    \label{tab:bioasqstats}
\end{table}

Given the demonstrated ability of LLMs to generate coherent and highly readable text, in this paper we want to test their performance in returning ``ideal'' answers using the benchmark from BioASQ 11b.

Until recently, there has been no published work that uses LLM and prompt engineering for answering biomedical questions from BioASQ. Common approaches to return the ``ideal'' answers used LLMs such as variants of BERT \cite{Devlin:2018}, but without prompt engineering. Here we briefly explain the system by \citet{Molla2022}, since it was used for some of the experiments reported here.\footnote{Public code of \citet{Molla2022} is available at \url{https://github.com/dmollaaliod/bioasq10b-public}} The system implemented query-focused extractive summarisation via text classification. The classifier used DistilBERT to obtain the word embeddings of the candidate snippet, followed by average pooling, and the final classification layer. DistilBERT used the question and candidate snippet as input, and the classification layer incorporated the snippet position. The top $n$ snippets were then returned as the ideal answer. This system produced competitive results in BioASQ 10b.\footnote{The code of top systems were not readily available when we conducted the experiments of this paper.}

At the time of writing this paper, the proceedings of BioASQ had just been released and several systems reported on the use of LLM and prompt engineering \cite[for example]{Ateia2023,Hsueh2023}. The top system by \citet{Hsueh2023} incorporated chain-of-thought and attempted the joint task of providing the exact answer and the ideal answer.

Besides BioASQ, but still within the medical domain, LLM and prompting engineering has been used to answer medical exam questions, including multiple-choice \cite{Lievin2023,Nori2023,Singhal2023} and long-form answers \cite{Singhal2023}. The questions used in BioASQ are not exam questions. Instead, they have been designed to simulate real information needs by biomedical experts \cite{Krithara2023}, and Figure~\ref{fig:example} is one such example. It is therefore interesting to know how prompt engineering for LLMs would perform with these questions.

Recently, \citet{Singhal2023} released a LLM that focuses on the biomedical domain. LLMs like this will presumably improve the quality of the answers to biomedical questions but, given time and resource constraints, they have not been used in the present study.

\section{Prompt Engineering}

In this section we detail the prompts used in our experiments with GPT-3.5. We used two types of prompts: without context, and with context.

\subsection{Prompts without context}

The first set of experiments explored the use of zero-shot and few-shot prompts that did not incorporate relevant snippets as context. Given the nature of the task, question answering of biomedical questions, the results are not expected to be of high quality; these experiments are baselines to test the potential improvement of prompt engineering.

As a trivial baseline, a \textbf{Zero-shot} variant simply used the biomedical question as a prompt, without any additional information. 

A \textbf{Few-shot} variant includes a short introductory text, plus the last $n=10$ question-answer pairs from the BioASQ 11b training data (Figure~\ref{fig:fewshotprompt}), before introducing the question to ask. By including these question-answer pairs, the system has access to examples that guide it in the process of generating the answer.

\begin{figure}
\centering
\fbox{
\begin{minipage}{7cm}
\em
Answer this biomedical question. Write the answer as the ideal answer given to a medical practitioner.\\
\\
Q: <question>\\
Q type: <qtype>\\
A: <answer>\\
\\
<9 more samples>\\
\\
Q: <question>\\
Q type: <qtype>\\
A: 
\end{minipage}
}
    \caption{Few-shot prompt using no context and $n=10$ training data samples.}
    \label{fig:fewshotprompt}
\end{figure}

\subsection{Retrieval Augmented Generation}

The second set of experiments used RAG that incorporated context in the prompt. Given that the BioASQ test data includes, besides the question and question type, a list of relevant snippets, we inserted these snippets. These snippets have been manually curated by the BioASQ annotators, who used retrieval tools during their annotation process \cite{Krithara2023}. These snippets, therefore, represent the ideal output of retrieval systems. Our prompt instructed the system to find the answer in the provided snippets (Figures~\ref{fig:contextprompt}--\ref{fig:bioasq10bfewshot}). 

In a \textbf{Zero-shot} setting, we experimented with two variants. In the \textbf{Snippets} variant (Figure~\ref{fig:contextprompt}), we inserted all the snippets as a list. In the \textbf{Extract} variant (Figure~\ref{fig:bioasq10bprompt}), we inserted an extract of the snippets. The extract is the unedited output of the DistilBERT QA system introduced by \citet{Molla2022}, and briefly described in Section~\ref{sec:related}.

\begin{figure}
\centering
\fbox{
\begin{minipage}{7cm}
\em
Answer the biomedical question as truthfully as possible using the provided list of snippets. Write the answer as the ideal answer given to a medical practitioner.\\
\\
Snippets:\\
\\
- <snippet 1>\\
\\
- <more snippets>\\
\\
Q: <question>\\
A: 
\end{minipage}
}
    \caption{Zero-shot prompt using all the relevant snippets as context.}
    \label{fig:contextprompt}
\end{figure}

\begin{figure}
\centering
\fbox{
\begin{minipage}{7cm}
\em
Answer the biomedical question as truthfully as possible using the provided text. Write the answer as the ideal answer given to a medical practitioner.\\
\\
Text:\\
\\
<extractive summary>\\
\\
Q: <question>\\
A: 
\end{minipage}
}
    \caption{Zero-shot prompt using an extractive summary of the relevant snippets as context.}
    \label{fig:bioasq10bprompt}
\end{figure}

\begin{figure}
    \centering
    \fbox{
    \begin{minipage}{7cm}
    \em
    Answer the biomedical question as truthfully as possible using the provided text. Write the answer as the ideal answer given to a medical practitioner.\\
    \\
    Text: <extractive summary>\\
    Q: <question>\\
    Q type: <qtype>\\
    A: <answer>\\
    \\
    <9 more samples>\\
    \\
    Text: <extractive summary>\\
    Q: <question>\\
    Q type: <qtype>\\
    A: 
    \end{minipage}
    }
        \caption{Few-shot prompt using an extractive summary of the relevant snippets as context.}
        \label{fig:bioasq10bfewshot}
    \end{figure}
    
A \textbf{Few-shot} variant (Figure~\ref{fig:bioasq10bfewshot}) included 10 samples of question, extractive summary, and target answer, to guide the system. The question and target answer are taken from the training data provided by BioASQ.\footnote{In particular, we used the annotated test data of BioASQ 10b, batch 2. This is a subset of the training data available for BioASQ 11b.} The extractive summary is the output of the system by \citet{Molla2022}. To make sure that the selected samples were useful to the specific question, we selected the last $n=10$ samples from the same question type (``yes/no'', ``summary'', ``factoid'', ``list'') 

\section{Results}

\begin{figure}
    \centering
    \begin{tikzpicture}
        \begin{axis}[
            xlabel={\footnotesize Batch},
            ylabel={\footnotesize Average Human},
            ymin=3,
            ymax=5,
            xtick={1,2,3,4},
            ticklabel style={font=\footnotesize},
            width=7.8cm,
            legend pos=south east,
            legend style={font=\tiny}
        ]
        \addplot coordinates {
        (1,4.7525)
        (2,4.8125)
        (3,4.8375)
        (4,4.8400)
        };
        \addplot coordinates {
        (1,4.1850)
        (2,4.3250)
        (3,4.3000)
        (4,4.3025)
        };
        \addplot coordinates {
        (1,4.0175)
        (2,4.3375)
        (3,4.2850)
        };
        \addplot coordinates {
        (1,4.1550)
        (2,4.5475)
        (3,4.4000)
        };
        \addplot coordinates {
        (1,4.6350)
        (2,4.5400)
        (3,4.7900)
        (4,4.715)
        };
        \addplot coordinates {
        (4,4.7575)
        };
        \addplot coordinates {
        (4,4.4575)
        };
        \legend{Top,Extract,Zero-shot No context (batches 1 to 3),Few-shot No context,Zero-shot Snippets,Zero-shot Extract (batch 4 only),Few-shot Extract (batch 4 only)}
        \end{axis}
    \end{tikzpicture}
    \caption{Plot of the experiment results, on a human evaluation scale that ranges from 1 to 5. The output of the ``Extract'' system was used as context for the systems ``Zero-shot Extract'' and ``Few-shot Extract''.}
    \label{fig:bioasqHumanResults}
\end{figure}
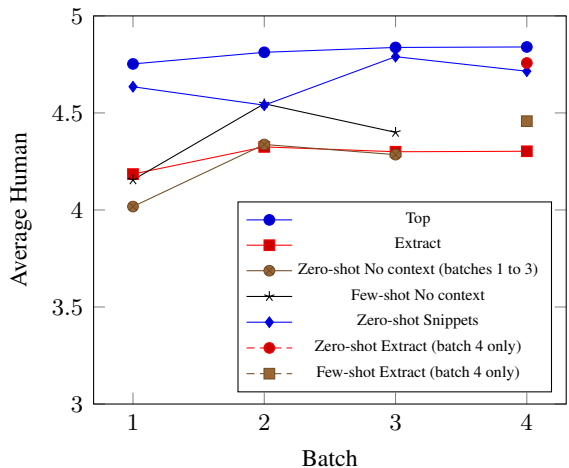

Figure~\ref{fig:bioasqHumanResults} plots the results of our experiments, using the test data of % BioASQ 10b batch 1, and 
BioASQ 11b, batches 1 to 4. These results are based on the average of 4 human evaluation criteria conducted by the organisers of BioASQ.\@ Refer to Appendix~\ref{appendix:results} for further details, including an automatic evaluation. For comparison, we also include the results of the top system, and the system developed by \citet{Molla2022}, named ``Extract'' in the figure.
%
% \addplot[color=red,mark=x] coordinates {
%
Appendix~\ref{appendix:answers} shows the generated answers for a selection of questions.

We observe the following:

\paragraph{Large Language Models improve over an extractive summariser.} With the prompts that used the output of an extractive summariser as context, the results were better than the results of the original extractive summariser.\footnote{The option of using the extractive summariser was made relatively late in the pipeline, and it could be tested in the last batch (batch 4) only.} As shown in the example outputs of Appendix~\ref{appendix:answers}, this is because the LLM can remove the irrelevant parts of the extract, and may use its background knowledge when the answer was not found in the extract. 

\paragraph{The best prompt uses retrieval-augmented context.} We observe that the best system uses the BioASQ list of snippets as context. Using the output of an extractive summariser does not appear to improve much. This suggests that it might not be necessary to do complex filtering of the initial snippets, since doing so may risk removing text that contains the answer, and the LLMs are capable of ignoring irrelevant snippets.

\paragraph{The results improve over approaches not using generative LLMs.} The results of our runs using context are better than those of other participating systems in 3 out of 4 batches of BioASQ 11b, except for \cite{Hsueh2023} who used GPT-4. This demonstrates the power of prompt engineering for this task.

Further work is required to study the impact of fine details of wording used in the prompts. Tools like PromptIDE (\citet{Strobelt2023}) may be useful for this.

Table~\ref{tab:parameters} details the parameters used in our runs, to facilitate reproducibility.

\begin{table}[h]
    \centering
    \begin{tabular}{ll}
       \bf Model  &  text-davinci-003\\
       \bf Temperature & 0 \\
       \bf Max Tokens & 200 \\
       \bf Top p & 1 \\
       \bf Frequency Penalty & 0.0\\
       \bf Presence Penalty & 0.0 
    \end{tabular}
    \caption{Parameters used in the runs. The runs used the Open AI API.}
    \label{tab:parameters}
\end{table}

\section{Conclusions}

In this paper, we have experimented with the use of prompting engineering for GPT-3.5. We observe that a retrieval-augmented prompt that uses relevant snippets as context improves over variants that do not provide context. We also observe that the system outperforms almost all other systems participating in BioASQ 11b.

Given that the designed prompts were relatively simple, this indicates that using LLMs and prompt engineering may represent a paradigm shift for biomedical QA. 

\section*{Limitations}

The relevant snippets have been curated by the BioASQ annotators, and consequently they would probably be of better quality than automatically-generated snippets. This means that the results of the prompts using context are probably an upper bound. Given the relatively large difference of results between versions with and without context, we presume that the general conclusions will still hold even when using automatically retrieved snippets. This is a task for future research. Note, incidentally, that our system outperformed most other systems in BioASQ 11b, where all systems were allowed to use these curated snippets.

\section*{Ethics Statement}

Since the approach presented here uses LLMs, the answers generated by this system may contain hallucinations and incorrect information. For this reason, we do not recommend the use of this system by people without biomedical training. The envisaged application scenario of this is a help tool to the medical practitioner or researcher.

The information used to obtain the answers is from public biomedical research papers from PubMed, and they are not likely to contain private or sensitive information. However, due to the nature of the original texts, they may contain details of surgical procedures that may be confronting to the untrained reader.

% \section*{Acknowledgements}

% \emph{(will be completed in the camera-ready version)}

\bibliography{anthology,custom}
\bibliographystyle{acl_natbib}

\appendix

\section{Results}
\label{appendix:results}

Table~\ref{tab:humanEvaluation} details the results of the human evaluation performed by the organisers of BioASQ, and Figure~\ref{fig:bioasqHumanResults} plots the results graphically.  The experiments are conducted on the test data of % BioASQ 10b batch 1, and 
BioASQ 11b, batches 1 to 4.
\begin{table*}[h]
    \centering
    \begin{tabular}{lllccccc}
    \toprule
      & & &&  \multicolumn{4}{c}{BioASQ 11b}\\
     System   && Context       && Batch 1 & Batch 2 & Batch 3 & Batch 4 \\
    \cmidrule{1-1} \cmidrule{3-3} \cmidrule{5-8}
    \em Top  &&                && 4.752   & 4.812   & 4.837   & 4.840\\
    \em Extract &&             && 4.185   & 4.325   & 4.300   & 4.302\\
    \cmidrule{1-1} \cmidrule{3-3} \cmidrule{5-8}
     \em Zero-shot && None     && 4.017   & 4.337   & 4.285   &\\
     \em Few-shot && None      && 4.155   & \textbf{4.547}   & 4.400   &\\
     \em Zero-shot && Snippets && \textbf{4.635}   & 4.540   & \textbf{4.790}   & 4.415\\
     \em Zero-shot && Extract  &&         &         &         & \textbf{4.757}\\
     \em Few-shot && Extract   &&         &         &         & 4.457\\
     \bottomrule
    \end{tabular}
    \caption{Human evaluation. The output of the system labelled ``Extract'' (second row of the table) was used as the context in the last two rows of the table. The system labelled ``Top'' indicates the best system participating in BioASQ 11b, besides our runs. Numbers in boldface indicate that our run outperformed the ``Top'' system. Metric values range from 1 to 5.}
    \label{tab:humanEvaluation}
\end{table*}
These results have been obtained after averaging the 4 evaluation criteria described by \cite{Tsatsaronis:2015} and reproduced below.

\begin{itemize}
    \item \textbf{Information recall (IR)} (1-5) -- All the necessary information is in the generated summary.
    \item \textbf{Information precision (IP)} (1-5) -- No irrelevant information is generated.
    \item \textbf{Information repetition (IRep)} (1-5) -- The generated summary does not repeat the same information multiple times.
    \item \textbf{Readability (Read)} (1-5) -- The generated summary is easily readable and fluent.
\end{itemize}

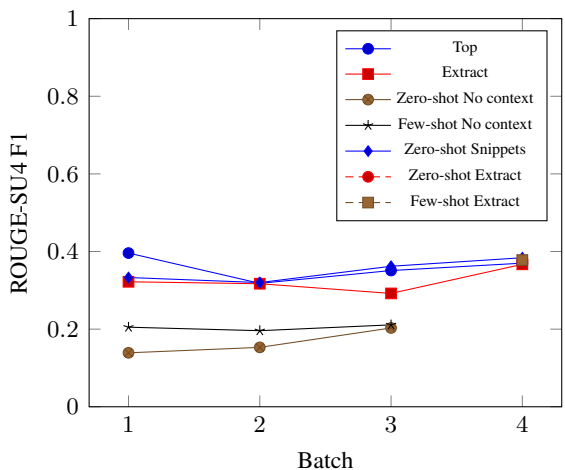
\begin{figure}
    \centering
    \begin{tikzpicture}
        \begin{axis}[
            xlabel={\footnotesize Batch},
            ylabel={\footnotesize ROUGE-SU4 F1},
            ymin=0,
            ymax=1,
            xtick={1,2,3,4},
            ticklabel style={font=\footnotesize},
            width=7.8cm,
            legend pos=north east,
            legend style={font=\tiny}
        ]
        \addplot coordinates {
        (1,0.396)
        (2,0.318)
        (3,0.351)
        (4,0.370)
        };
        \addplot coordinates {
        (1,0.322)
        (2,0.317)
        (3,0.292)
        (4,0.367)
        };
        \addplot coordinates {
        (1,0.139)
        (2,0.153)
        (3,0.203)
        };
        \addplot coordinates {
        (1,0.205)
        (2,0.196)
        (3,0.211)
        };
        \addplot coordinates {
        (1,0.333)
        (2,0.320)
        (3,0.362)
        (4,0.384)
        };
        \addplot coordinates {
        (4,0.377)
        };
        \addplot coordinates {
        (4,0.378)
        };
        \legend{Top,Extract,Zero-shot No context,Few-shot No context,Zero-shot Snippets,Zero-shot Extract,Few-shot Extract}
        \end{axis}
    \end{tikzpicture}
    \caption{Plot of the experiment results. The output of the ``Extract'' system was used as context for the systems ``Zero-shot Extract'' and ``Few-shot Extract''.}
    \label{fig:bioasqResults}
\end{figure}

For completness, and to facilitate comparison with systems developed in the future and which might not have human evaluations, Table~\ref{tab:bioasqResults} details the results of the automatic evaluation of the experiments described in this paper, using the test data of % BioASQ 10b batch 1, and 
BioASQ 11b, batches 1 to 4. The evaluation metric is ROUGE-SU4 F1 between the generated answer and the golden answer.\footnote{A question may have several golden answers. When that happens, the maximum ROUGE-SU4 F1 score is used.} For comparison, we also include the results of the top system (outside our runs), and the system developed by \citet{Molla2022}, named ``Extract'' in the figure.
\begin{table*}[h]
    \centering
    \begin{tabular}{lllccccc}
    \toprule
      & & &&  \multicolumn{4}{c}{BioASQ 11b}\\
     System   && Context       && Batch 1 & Batch 2 & Batch 3 & Batch 4 \\
    \cmidrule{1-1} \cmidrule{3-3} \cmidrule{5-8}
    \em Top  &&                && 0.396   & 0.318   & 0.351   & 0.370\\
    \em Extract &&             && 0.322   & 0.317   & 0.292   & 0.367\\
    \cmidrule{1-1} \cmidrule{3-3} \cmidrule{5-8}
     \em Zero-shot && None     && 0.139   & 0.153   & 0.203   &\\
     \em Few-shot && None      && 0.205   & 0.196   & 0.211   &\\
     \em Zero-shot && Snippets && \textbf{0.333}   & \emph{\textbf{0.320}}   & \emph{\textbf{0.362}}   & \emph{\textbf{0.384}}\\
     \em Zero-shot && Extract  &&         &         &         & 0.377\\
     \em Few-shot && Extract   &&         &         &         & 0.378\\
     \bottomrule
    \end{tabular}
    \caption{Results of the experiments. The output of the system labelled ``Extract'' (second row of the table) was used as the context in the last two rows of the table. The system labelled ``Top'' indicates the best system participating in BioASQ 11b, besides our runs. Numbers in italics indicate that our run outperformed the ``top'' system. Numbers in boldface indicate the best evaluation result within our runs. Metric: ROUGE-SU4 F1.}
    \label{tab:bioasqResults}
\end{table*}
Figure~\ref{fig:bioasqResults} plots the results graphically. 

% \addplot[color=red,mark=x] coordinates {
%

Note that the ROUGE results may appear low when compared with other summarisation tasks. This is a characteristic of BioASQ, where ROUGE results are usually low. BioASQ conducts human evaluations, which can reach high scores in the top runs. The human evaluation results, on the other hand, are very high, with values over~4 in a range of~1 to~5.

\section{Sample Outputs}
\label{appendix:answers}

This appendix shows the answer generated by the systems in a sample of every question type of batch 1 (batch 4 for runs that were not submitted to batch 1). The target answer will also be included in the camera ready version if it becomes available at the time of writing.

In the answers shown below, the system labelled ``Extract'' is the output of the extractive summariser that was used as context for the zero-shot and few-shot systems labelled with ``\ldots extractive summary as context''.

\paragraph{Batch 1:}
\begin{description}
  \item[yes/no]
  \begin{description}
      \item[Q:] \emph{Can losartan reduce brain atrophy in Alzheimer's disease?}
      \item[Zero shot, no context:] \emph{No, losartan is not known to reduce brain atrophy in Alzheimer's disease. However, there are medications that have been shown to slow the progression of Alzheimer's disease, such as donepezil, rivastigmine, and memantine.}
      \item[Few-shot, no context:] \emph{Yes, losartan has been shown to reduce brain atrophy in patients with Alzheimer's disease.}
      \item[Zero-shot, snippets as context:] \emph{No, 12 months of treatment with losartan was not effective in reducing the rate of brain atrophy in individuals with clinically diagnosed mild-to-moderate Alzheimer's disease.}
      \item[Extract:] \emph{INTERPRETATION: 12 months of treatment with losartan was well tolerated but was not effective in reducing the rate of brain atrophy in individuals with clinically diagnosed mild-to-moderate Alzheimer's disease.}
  \end{description}
  \item[factoid]
  \begin{description}
      \item[Q:] \emph{What is CHARMS with respect to medical review of predictive modeling?}
      \item[Zero shot, no context:] \emph{CHARMS (Clinical Health Analytics Review and Modeling System) is a software platform designed to facilitate the medical review of predictive modeling. It provides a comprehensive suite of tools to help healthcare organizations evaluate and validate predictive models, including data visualization, model comparison, and model performance metrics. CHARMS also provides a secure environment for sharing and collaborating on predictive models.}
      \item[Few-shot, no context:] \emph{CHARMS (Clinical Hierarchy Attribute Reduction for Model Simplification) is a method for medical review of predictive modeling that uses a hierarchical structure of clinical attributes to reduce the complexity of a predictive model.}
      \item[Zero-shot, snippets as context:] \emph{CHARMS stands for Checklist for Critical Appraisal and Data Extraction for Systematic Reviews of Prediction Modeling Studies. It is a tool used to assess the quality of systematic reviews of predictive modeling studies in the medical field.}
      \item[Extract:] \emph{CHecklist for critical Appraisal and data extraction for systematic Reviews of prediction Modelling Studies (CHARMS). We will extract data based on the Checklist for\u00a0Critical Appraisal and Data Extraction for Systematic Reviews of Prediction Modeling Studies (CHARMS),}
  \end{description}
  \item[list]
  \begin{description}
      \item[Q:] \emph{Which splicing factors have been associated with alternative splicing in PLN R14del hearts?}
      \item[Zero shot, no context:] \emph{The splicing factors that have been associated with alternative splicing in PLN R14del hearts include: SRPK1, SRPK2, SRP14, SRP19, and hnRNPA2/B1.}
      \item[Few-shot, no context:] \emph{Splicing factors that have been associated with alternative splicing in PLN R14del hearts include hnRNPA2/B1, hnRNPC, PTBP1, and SRSF1.}
      \item[Zero-shot, snippets as context:] \emph{Bioinformatical analysis has pointed to the tissue-specific splicing factors Srrm4 and Nova1 as likely upstream regulators of the observed splicing changes in the PLN-R14del cardiomyocytes.}
      \item[Extract:] \emph{Our work suggests that an intricate interplay of programs controlling gene expression levels and AS is fundamental to organ development, especially for the brain and heart. Bioinformatical analysis pointed to the tissue-specific splicing factors Srrm4 and Nova1 as likely upstream regulators of the observed splicing changes in the PLN-R14del cardiomyocytes.}
  \end{description}
  \item[summary]
  \begin{description}
      \item[Q:] \emph{Which are the targets of Tirzepatide?}
      \item[Zero shot, no context:] \emph{Tirzepatide is an investigational drug being studied for the treatment of type 2 diabetes. It is a dual agonist of the GLP-1 and glucagon receptors, which are hormones that regulate glucose levels in the body. The primary targets of Tirzepatide are to improve glycemic control, reduce body weight, and reduce cardiovascular risk factors.}
      \item[Few-shot, no context:] \emph{Tirzepatide is a dual-acting GLP-1/glucagon receptor agonist that targets both GLP-1 and glucagon receptors. It is being investigated for the treatment of type 2 diabetes.}
      \item[Zero-shot, snippets as context:] \emph{Tirzepatide is a dual GLP-1R/GIPR agonist, targeting both the glucagon-like peptide 1 (GLP-1) and glucose-dependent insulinotropic polypeptide (GIP) receptors. It is a new antidiabetic drug with potential metabolic activity in the treatment of type 2 diabetes.}
      \item[Extract:] \emph{Designing a Dual GLP-1R/GIPR Agonist from Tirzepatide: Comparing Residues Between Tirzepatide, GLP-1, and GIP. he efficacy and safety of tirzepatide, a novel glucose-dependent insulinotropic polypeptide and glucagon-like peptide-1 receptor agonist, in people with obesity are not known. Tirzepatide - a dual GIP/GLP-1 receptor agonist - a new antidiabetic drug with potential metabolic activity in the treatment of type 2 diabetes. The incretin hormones glucagon-like peptide 1 (GLP-1) and glucose-dependent insulinotropic polypeptide (GIP) are responsible for up to 65\% of postprandial insulin secretion. Tirzepatide, developed by Eli Lilly, is a dual GIP/GLP-1 receptor agonist in the form of a synthetic linear peptide; its acylation technology allows it to bind to albumin, thus making it possible to dose the drug once a week.}
  \end{description}
\end{description}

\paragraph{Batch 4:}

\begin{description}
    \item[yes/no]
    \begin{description}
        \item[Q:] \emph{Is Functional Neurological Disorder (FND) classified as a psychosomatic disorder?}
        \item[Zero-shot, snippets as context:] \emph{Functional Neurological Disorder (FND) is not classified as a psychosomatic disorder, but rather as a neurological disorder. It is characterized by impairments in brain networks leading to distressing motor, sensory and/or cognitive symptoms that demonstrate positive clinical signs on examination incongruent with other conditions. It is also associated with maladaptive schemas and trauma-related maladaptive schemas, as well as aberrant communication between neural networks.}
        \item[Zero-shot, extractive summary as context:] \emph{Yes, Functional Neurological Disorder (FND) is classified as a psychosomatic disorder. It is a neurological disorder that is caused by psychological factors, such as stress, anxiety, or trauma. It can also be associated with somatic symptom disorder in Parkinson's disease.}
        \item[Few-shot, extractive summary as context:] \emph{Yes, Functional Neurological Disorder (FND) is classified as a psychosomatic disorder.}
        \item[Extract:] \emph{Challenges to the Diagnosis of Functional Neurological Disorder: Feigning, Intentionality, and Responsibility. Functional neurological disorder and somatic symptom disorder in Parkinson's disease.}
    \end{description}
    \item[factoid]
    \begin{description}
        \item[Q:] \emph{Which amino acid in implicated in the Blue diaper syndrome?}
        \item[Zero-shot, snippets as context:] \emph{Tryptophan is implicated in Blue diaper syndrome (BDS).}
        \item[Zero-shot, extractive summary as context:] \emph{No specific amino acid has been implicated in Blue diaper syndrome.}
        \item[Few-shot, extractive summary as context:] \emph{Blue diaper syndrome (BDS) is caused by a mutation in the SLC6A19 gene, which encodes a transporter protein for the amino acid cystine.}
        \item[Extract:] \emph{Blue diaper syndrome (BDS) (Online Mendelian Inheritance in Man number 211000) is an extremely rare disorder that was first described in 1964. The characteristic finding is a bluish discoloration of urine spots in the diapers of affected infants.}
    \end{description}
    \item[list]
    \begin{description}
        \item[Q:] \emph{What were the primary endpoints of the phase I/II double-blind, comparator-controlled trial of RT001 in Friedreich\u0027s ataxia patients?}
        \item[Zero-shot, snippets as context:] \emph{The primary endpoints of the phase I/II double-blind, comparator-controlled trial of RT001 in Friedreich's ataxia patients were safety, tolerability, and pharmacokinetic analysis.}
        \item[Zero-shot, extractive summary as context:] \emph{The primary endpoints of the phase I/II double-blind, comparator-controlled trial of RT001 in Friedreich's ataxia patients were safety, tolerability, and pharmacokinetic analysis.}
        \item[Few-shot, extractive summary as context:] \emph{The primary endpoints of the phase I/II double-blind, comparator-controlled trial of RT001 in Friedreich's ataxia patients were safety, tolerability, and pharmacokinetic analysis.}
        \item[Extract:] \emph{To evaluate the safety, pharmacokinetics, and preliminary efficacy of RT001 in Friedreich's ataxia patients The primary endpoints were safety, tolerability, and pharmacokinetic analysis.}
    \end{description}
    \item[summary]
    \begin{description}
        \item[Q:] \emph{Explain the function of SOX9}
        \item[Zero-shot, snippets as context:] \emph{SOX9 is a transcription factor with critical roles in many biological processes, including neurodevelopment, chondrocyte differentiation, testis formation, thyroid differentiation, and cancer progression. It has been shown to be upregulated in many tumours and to mediate immunosuppressive tumour microenvironment by suppressing CD8+ T cell responses and modified macrophage function. Additionally, SOX9 has been found to be co-expressed with other genes in PCL of humans and mice, and to be regulated by enhancers and crosstalk between TSH, TGF\u03b2 and thyroid transcription factors.}
        \item[Zero-shot, extractive summary as context:] \emph{SOX9 is a transcription factor with critical roles during neurodevelopment, particularly within neural stem cells. It also plays a role in the differentiation of mesenchymal progenitor cells into chondrocytes. In addition, SOX9 has been found to be upregulated in many cancers, where it is thought to mediate an immunosuppressive tumour microenvironment by suppressing CD8+ T cell responses and modified macrophage function.}
        \item[Few-shot, extractive summary as context:] \emph{SOX9 is a transcription factor with critical roles during neurodevelopment, particularly within neural stem cells. It also plays a role in the differentiation of mesenchymal progenitor cells into chondrocytes. In cancer, SOX9 has been upregulated and is involved in mediating immunosuppressive tumour microenvironment by suppressing CD8+ T cell responses and modified macrophage function in gastric adenocarcinoma through the paracrine LIF factor.}
        \item[Extract:] \emph{Targeted knockout of the SOX9 gene, which plays a role in the differentiation of mesenchymal progenitor cells into chondrocytes,  SOX9 is a transcription factor with critical roles during neurodevelopment, particularly within neural stem cells. Epithelial SOX9 drives progression and metastases of gastric adenocarcinoma by promoting immunosuppressive tumour microenvironment. any cancers engage embryonic genes for rapid growth and evading the immune system. SOX9 has been upregulated in many tumours, yet the role of SOX9 in mediating immunosuppressive tumour microenvironment is unclear. Epithelial SOX9 is critical in suppressing CD8+ T cell responses and modified macrophage function in GAC through the paracrine LIF factor.}
    \end{description}
\end{description}

% This is a section in the appendix.

\end{document}